%% file: main.tex
\documentclass[11pt]{article}
\usepackage[margin=1.15in]{geometry}
\IfFileExists{newtxtext.sty}
  {\usepackage{newtxtext}}
  {\usepackage{mathptmx}}
\usepackage[numbers]{natbib}

\usepackage[T1]{fontenc}
\usepackage[utf8]{inputenc}
\usepackage{microtype}
\usepackage{upquote}
\usepackage{xparse}
\usepackage{xspace}

\usepackage{amsmath}
\usepackage{amssymb}
\IfFileExists{newtxmath.sty}{\usepackage{newtxmath}}{}

\usepackage{booktabs}
\usepackage{array}
\usepackage{longtable}
\usepackage{graphicx}
\usepackage[font=small]{caption}
\usepackage{xcolor}
\usepackage{listings}
\usepackage{fancybox}
\usepackage{pdfrender}
\usepackage{url}
\usepackage[colorlinks=true,
            linkcolor=black,
            citecolor=black,
            urlcolor=black,
            breaklinks=true]{hyperref}
\usepackage[capitalise,noabbrev]{cleveref}

\DeclareUnicodeCharacter{2759}{\rule[-0.15ex]{0.7pt}{1.6ex}}
\DeclareUnicodeCharacter{2717}{$\times$}

\definecolor{burgundy}{RGB}{128,0,32}
\definecolor{darkred}{rgb}{0.68,0.05,0.0}
\definecolor{blue(pigment)}{rgb}{0.2,0.2,0.6}
\definecolor{metagray}{gray}{0.45}
\definecolor{lstbg}{gray}{0.97}

\NewDocumentCommand{\mildbold}{O{0.2pt} O{black} m}{%
  {\pdfrender{
    TextRenderingMode=FillStroke,
    LineWidth=#1,
    FillColor=#2,
    StrokeColor=#2
  }#3}%
}

\ExplSyntaxOn
\tl_new:N  \g_ul_style_tl
\tl_new:N  \g_ul_color_tl
\dim_new:N \g_ul_width_dim
\dim_new:N \g_ul_gap_dim
\dim_new:N \g_ul_dash_dim
\dim_new:N \g_ul_above_dim
\dim_new:N \g_ul_below_dim

\keys_define:nn { uline }
{
  style .tl_gset:N  = \g_ul_style_tl,
  color .tl_gset:N  = \g_ul_color_tl,
  width .dim_gset:N = \g_ul_width_dim,
  gap   .dim_gset:N = \g_ul_gap_dim,
  dash  .dim_gset:N = \g_ul_dash_dim,
  above .dim_gset:N = \g_ul_above_dim,
  below .dim_gset:N = \g_ul_below_dim,
}

\cs_new_protected:Npn \LineParse #1
{
  \keys_set:nn { uline }
  {
    style=solid,
    color=burgundy,
    width=1pt,
    gap=2pt,
    dash=4pt,
    above=1.5pt,
    below=1.5pt,
    #1
  }
  \xdef\LineStyle{\tl_use:N \g_ul_style_tl}
  \xdef\LineColor{\tl_use:N \g_ul_color_tl}
  \xdef\LineWidth{\dim_use:N \g_ul_width_dim}
  \xdef\LineGap{\dim_use:N \g_ul_gap_dim}
  \xdef\LineDash{\dim_use:N \g_ul_dash_dim}
  \xdef\LineAbove{\dim_use:N \g_ul_above_dim}
  \xdef\LineBelow{\dim_use:N \g_ul_below_dim}
}
\ExplSyntaxOff

\def\Line#1#2{%
  \csname Line@#1\endcsname{#2}%
}

\expandafter\def\csname Line@text\endcsname#1{%
  \begingroup
  \LineParse{#1}%
  \par\vspace{\LineAbove}%
  \noindent
  \csname TextLine@\LineStyle\endcsname
  \par\vspace{\LineBelow}%
  \endgroup
}

\expandafter\def\csname TextLine@dashed\endcsname{%
  \leavevmode
  \color{\LineColor}%
  \leaders
    \hbox{\rule{\LineDash}{\LineWidth}\hspace{\LineGap}}%
  \hfill
  \kern0pt
}

\expandafter\def\csname Line@table\endcsname#1{%
  \noalign{\LineParse{#1}}%
  \csname TableLine@\LineStyle\endcsname
}

\expandafter\def\csname TableLine@solid\endcsname{%
  \noalign{%
    \vskip\LineAbove
    {\color{\LineColor}\hrule height \LineWidth}%
    \vskip\LineBelow
  }%
}

\hypersetup{
  allcolors=burgundy,
  pdftitle={FedBrew: Paper-Faithful Federated Learning: Scientific Checks
            for Practitioner and Agent Development},
  pdfauthor={Ali Dadras and Alp Yurtsever},
  pdfsubject={Federated learning research software},
}

\newcommand{\fedbrew}{\mildbold{FedBrew}\xspace}
\newcommand{\fbrepo}{github.com/DadrasAli/fedbrew}
\newcommand{\fbrepourl}{https://github.com/DadrasAli/fedbrew}
\newcommand{\fbsubtitle}{Paper-Faithful Federated Learning:\\
Scientific Checks for Practitioner and Agent Development}
\newcommand{\opt}[1]{-{}-#1}

\newlength{\cmdRuleWidth}
\newlength{\cmdRuleGap}
\newsavebox{\fbcmdbox}

\newenvironment{commandblock}{%
  \par\addvspace{0.35em}%
  \begin{lrbox}{\fbcmdbox}%
  \begin{minipage}{\dimexpr\linewidth-\cmdRuleWidth-\cmdRuleGap\relax}%
  \ttfamily\footnotesize
  \setlength{\parindent}{0pt}%
  \setlength{\parskip}{0.42em}%
}{%
  \end{minipage}%
  \end{lrbox}%
  \noindent
  \textcolor{burgundy}{%
    \rule[-\dp\fbcmdbox]{\cmdRuleWidth}{\dimexpr\ht\fbcmdbox+\dp\fbcmdbox\relax}}%
  \hspace{\cmdRuleGap}\usebox{\fbcmdbox}%
  \par\addvspace{0.35em}%
}
\newcommand{\cmdline}[1]{\textcolor{metagray}{\$}\hspace{0.65em}#1\par}

\begin{document}
\thispagestyle{empty}

\begin{center}
  \includegraphics[height=4.5em]{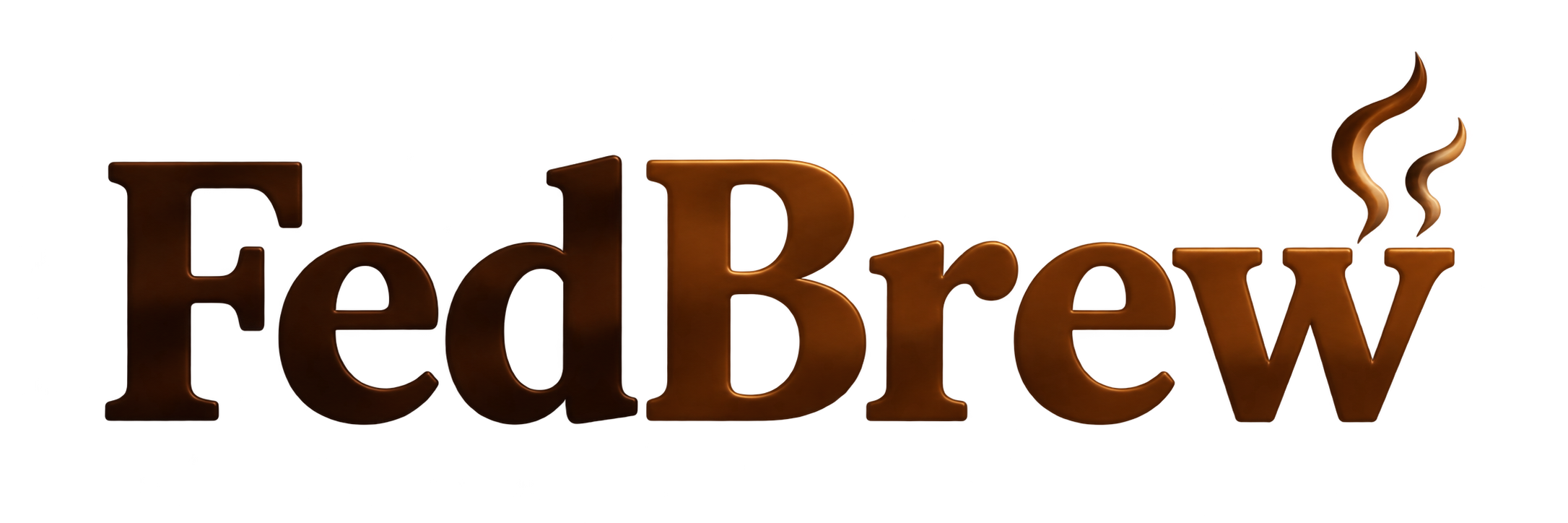}\par
  \vspace{0.8em}
  {\large\bfseries\fbsubtitle\par}

  \vspace{1.10em}
  {\large Ali Dadras \qquad Alp Yurtsever\par}
  \vspace{0.40em}
  {\normalsize Department of Mathematics and Mathematical Statistics\\
   Ume\aa{} University\par}
  \vspace{0.30em}
  {\small\ttfamily ali.dadras@umu.se \quad alp.yurtsever@umu.se\par}

  \vspace{0.95em}
  {\normalsize\color{metagray}\href{\fbrepourl}{\fbrepo}\par}
\end{center}

\begin{abstract}
Federated learning experiments can produce reasonable-looking results even when the experiment implementation differs from the intended experimental protocol. Small errors in configuration, data partitioning, aggregation, randomness, evaluation, or checkpoint recovery may therefore affect a comparison without causing the program to fail. We present FedBrew, a research codebase for controlled and reproducible federated learning experiments. Its focus is controlled simulation with checks of specified conditions of the intended experimental protocol. FedBrew uses configuration checks, identifiable client data partitions, checkpoint recovery checks, self-describing run records, and development-time checks of scientific conditions. These checks assess, for example, whether client partitions are disjoint and exhaustive and aggregation weights match the prescribed rule; numerical and recovery checks examine non-finite client model states and interrupted runs. The same structure also provides clear extension points and automated checks for new algorithms, datasets, models, and tasks, whether they are added by researchers or AI coding agents. 
FedBrew therefore supports reproducible and inspectable FL experiments through checks of specified software and scientific conditions and the run records associated with reported results.
\end{abstract}

\Line{text}{
  style=dashed,
  color=burgundy,
  width=1.5pt,
  above=25pt,
  below=-5pt,
  dash=5pt,
  gap=3pt
}
\section*{Quick Start}
\vspace{-0.35em}
{\small
Run the built-in synthetic FedAvg example to exercise the complete workflow \cite{mcmahan2017communication}. No
GPU is required. Download the \href{\fbrepourl}{repository}, open its folder,
and run:}

\begin{commandblock}
\cmdline{pip install -e .}
\cmdline{fedbrew generate synthetic}
\cmdline{fedbrew run synthetic/fedavg}
\end{commandblock}

\vspace{0.25em}
{\small Find the results in
{\footnotesize\texttt{outputs/synthetic/fedavg/}}}
\Line{text}{
  style=dashed,
  color=burgundy,
  width=1.5pt,
  above=5pt,
  below=5pt,
  dash=5pt,
  gap=3pt
}

\input{sections/01-introduction}
\input{sections/02-system-design}
\input{sections/03-contracts}
\input{sections/04-agent-ready}
\input{sections/05-conclusion}

\section*{Acknowledgments}
This work was partially supported by the Wallenberg AI, Autonomous Systems and Software Program (WASP) funded by the Knut and Alice Wallenberg Foundation. This work was also supported by the Swedish Research Council (registration number 2023-05476).

\setlength{\bibsep}{0.5em}
\bibliographystyle{unsrtnat}
\bibliography{refs}

\end{document}

%% file: sections/01-introduction.tex
\section{Introduction}
\label{sec:intro}

Federated learning (FL) trains a shared model across multiple clients
that keep their data locally. In a typical communication round, a server sends
the current model to selected clients, the clients perform local training, and
the server aggregates their results to update the shared model.
Here, the \emph{experimental protocol} is the scientific specification of how
an FL experiment is intended to be carried out and evaluated. It includes
the algorithm and its update rules, client data partitions and sampling,
randomness, evaluation, and checkpoint recovery. It describes the procedure
the researcher intends to study and report, whether in a paper or an experiment
plan, rather than the code that happens to execute. We use \emph{protocol}
in this scientific sense throughout.

The \emph{experiment implementation} is the software that realizes this
protocol, from data preparation and client selection to local training,
aggregation, evaluation, and recording results. A run can finish without
errors even if some part of this computation differs from the intended
protocol. For example, in a classification experiment using sample-size-weighted
FedAvg, the protocol may prescribe aggregation weights equal to the clients'
full training-split sizes. Weighting instead by the number of examples processed
in one local update changes the aggregation rule, even if the run completes
and produces plausible results. Rerunning the same code can reproduce
this deviation.

We distinguish \emph{computational reproducibility},
meaning that repeated executions under specified conditions produce consistent
results, from \emph{protocol fidelity}, meaning that the executed procedure
agrees with the experimental protocol the researcher intended to study.
An implementation may exhibit computational reproducibility without achieving protocol fidelity.

This becomes especially important when competing methods differ only modestly
in reported performance. Seed selection, data partitioning, hyperparameters,
evaluation rules, and recovery behavior (including whether an interrupted
run resumes consistently) can all substantially affect the comparison. In that regime,
keeping the experimental protocol controlled and inspectable is part of
establishing whether an \emph{observed difference comes from the algorithm itself.}

As research software evolves, new algorithms, datasets,
and experimental modes introduce interactions between components that
may behave correctly in isolation but incorrectly when combined.
These interactions make \emph{software compatibility} necessary:
components must obey the interfaces expected by the rest of the system.
For instance, a newly introduced client update rule must be documented
and work with the server strategies and tasks with which it is used.
Yet software compatibility alone does not establish protocol fidelity:
compatible components may still implement the wrong aggregation or
evaluation rule. Changes made by researchers and AI coding agents must
therefore be assessed on both grounds.

We use \mildbold[0.15pt][burgundy]{\emph{check}} to mean an automated procedure that examines a
specified condition on given inputs and reports whether it holds.
We distinguish \emph{software conditions}, such as recognized configuration
settings and supported server--client combinations, from
\emph{scientific conditions}, which are verifiable properties of the
intended experimental protocol. The prescribed FedAvg aggregation weights
above are a scientific condition; following a server's input and output
interfaces is a software condition. A check may run before training,
during execution, or when developers evaluate a change to the code. The word
\emph{check} refers to the same kind of automated verification regardless
of when it is performed.

For the FedAvg example, a check
performed when the implementation is developed or modified can compare
the server's aggregate for a prescribed set of client models with the
corresponding weighted average. This checks a specified scientific
condition; it is not necessarily repeated in every training round.
By contrast, rejecting an unknown configuration setting before training
checks a software condition, not the scientific aggregation rule.
Checks can detect violations of stated conditions within their scope,
but cannot determine which scientific conditions the researcher ought
to have specified. Unstated assumptions still require scientific judgment
and review.

In this paper, we ask how a federated learning research codebase can
remain scientifically precise, reproducible, maintainable, and safely
extensible by both practitioners and AI coding agents. We present
\fedbrew, a simulator for controlled federated-learning
experiments. \fedbrew uses shared interfaces to define how server strategies, client
update rules, and tasks interact. It checks configurations and supported
component combinations. It also provides identifiable data partitions,
recoverable runs, and checks of selected scientific conditions.
The checks make specified departures from the intended protocol more
visible. Distributed deployment and secure aggregation are outside the scope. 

This paper presents three aspects of
\fedbrew: (i) a modular FL simulator with common procedures
for client selection, local training, aggregation, evaluation, and recording
results; (ii) checks for selected software and scientific conditions
concerning configuration, data, aggregation, numerical behavior, and recovery;
and (iii) a workflow for introducing algorithms and other components 
while
checking their software compatibility and stated scientific conditions.
We describe these mechanisms and illustrate their checks with concrete
cases from the implementation.

\subsection{Related Work}
\label{sec:related-work}

Several FL frameworks support research experimentation with different
design emphases. \emph{fluke} provides a lightweight environment for
algorithm prototyping and reproducible simulation \cite{polato2026fluke}.
FLGo offers a plugin-based platform with reusable benchmarks, experiment
tracking, and analysis tools \cite{wang2023flgo}. FederatedScope uses an
event-driven architecture to support heterogeneous participant behaviors
and synchronous or asynchronous execution \cite{xie2023federatedscope}.
Flower, FedML, FedScale, and LEAF provide further FL execution and
benchmarking infrastructure
\cite{beutel2020flower,he2020fedml,lai2022fedscale,caldas2018leaf}.

Beyond FL frameworks, configuration and data dependencies have been
identified as sources of technical debt in machine-learning systems
\cite{sculley2015technicaldebt}. Design by contract
\cite{meyer1992contract}, property-based testing
\cite{claessen2000quickcheck}, and metamorphic testing
\cite{segura2016metamorphic} provide ways to express and examine
implementation properties; mutation testing assesses whether checks
detect deliberately introduced faults \cite{jia2011mutation}.
Research on scientific reproducibility also considers executable
checks for paper-code auditing \cite{peimani2026esc},
implementation requirements for automated reproduction
\cite{hu2026reproagent}, machine-readable descriptions of scientific
assumptions and validity domains \cite{menzo2026harness}, and
independently gradable reproduction obligations
\cite{starace2025paperbench}.

\fedbrew is a standalone FL research codebase organized around explicit
conditions of the intended experimental protocol, checks at the stages
where those conditions can be examined, and records that support the
inspection of reported results.

%% file: sections/02-system-design.tex
\section{\texorpdfstring{\fedbrew}{FedBrew} Design}
\label{sec:fedbrew-design}

\fedbrew provides checks for software conditions concerning configuration and interfaces as well as selected scientific conditions concerning data and algorithmic behavior. It also records information needed to inspect and resume experiments. 

This focus is reflected in the execution model. In each simulated communication round, \fedbrew selects clients, broadcasts the global model to the selected clients, performs their local updates, aggregates the results, evaluates the model, and records metrics and checkpoints. The round sequence itself is common to FL simulators; the distinction pursued here is to connect its stages to explicitly stated software and scientific conditions. Before training, configuration and interface checks reject specified invalid settings and component combinations; during aggregation, numerical checks reject non-finite client states; and development-time checks compare selected scientific conditions, such as those governing FedAvg aggregation, with reference calculations. The run record preserves the resolved configuration and, when available, the source-code revision and dataset-manifest hash for later inspection. These mechanisms target specified deviations from the intended experimental protocol rather than treating successful execution as proof of protocol fidelity.

\subsection{Architecture for Reliable Experimentation}

\fedbrew exposes six registries for server strategies, client update
rules, tasks, datasets, models, and data generators. Each registry maps
a component name to the function that constructs it. The training loop is assembled from the first five, while data
generation uses the sixth. 
Components depend on shared interfaces rather than on a particular algorithm or workload, keeping component-specific logic outside the
orchestration path. In particular, server strategies govern client selection and
aggregation, client update rules define local training, and task adapters provide
model, data, and evaluation behavior. The round loop coordinates these
components without embedding their method-specific implementations. These interfaces define software compatibility but do not, by
themselves, establish that the selected algorithm follows the scientific conditions specified in the intended experimental protocol.

\vspace{0.5em}

\noindent The following structural choices support a consistent experiment implementation and make selected conditions easier to check:

\begin{list}{{\color{darkred!90}\tiny$\blacksquare$}}{\leftmargin=1.5em}\vspace*{-5pt}
\setlength{\itemsep}{2pt}
\item 

Client results are folded into a streaming weighted aggregate as they
arrive rather than collected until the end of a round. The aggregation state therefore requires \(O(P)\) memory for \(P\) federated parameters, independent of the number of participating clients. This bound does not include client training or batched execution memory.

\item 

Data generation is separate from training. A generator writes the
per-client data files once and records how they form the partition in a
manifest; later runs reuse that manifest instead of generating a new split.
\fedbrew also records a hash of the manifest with the run, which identifies the partition description. For example, if the manifest
at the same path is replaced, a new run records a different manifest hash. Before training, it checks that the files named by
the manifest exist and that their structure agrees with what the manifest
declares. 

The manifest, its recorded hash, and the pre-training checks make
the generated partition identifiable and allow its declared file
structure to be checked against the referenced data files.

\item 

Built-in and external components enter through the same registries and
construction path. Each registry maps a component's config-file name to the
function that builds it. Extensions are loaded through an explicit
\texttt{register()} hook before configuration names are resolved; their declared
configuration keys are accepted only when that component is selected and are
passed to its construction function. 

New components can therefore use the standard command-line interface (CLI),
configuration checks, and artifact-recording pipeline without modifying
the core package. External server--client combinations may still require
additional checks of software compatibility and scientific conditions
provided by the extension author.

\item

Servers and clients exchange information through a small task-agnostic
request/result interface. Any
algorithm-specific auxiliary state is carried in a \emph{payload}, an extra
field attached to the shared request and result objects. SCAFFOLD
\cite{karimireddy2020scaffold} is one example: it corrects for client
drift. Its control variate -- a per-client correction added to the local
update -- travels inside that payload without the round loop inspecting it.
Control variates, optimizer moments, and similar auxiliary state can
therefore be introduced without changing the round loop or the shared
request/result objects.
In \fedbrew, the exchanged model states remain in memory rather
than being transferred over a network.

\item 

A \texttt{TaskAdapter} is the task-specific bridge between the generic federated orchestration and the task/model/data logic. It builds the task model and data loader, defines training and evaluation steps
and metrics, and specifies which model tensors are federated. It may also
define aggregation weights, which are used only by client update rules
that support the corresponding task hook.
This separation also accommodates different choices of federated model state. For example, in language-model fine-tuning with LoRA \cite{hu2021lora}, the task adapter can specify only the adapter tensors, rather than the entire model, as the federated state, while the round loop remains unchanged.
\end{list}

\Cref{tab:requirements} summarizes representative checks and the stages at which they are performed.

\begingroup
\small
\setlength{\tabcolsep}{4pt}
\renewcommand{\arraystretch}{1.0}
\begin{longtable}{
>{\raggedright\arraybackslash}m{0.20\textwidth}
>{\raggedright\arraybackslash}m{0.48\textwidth}
>{\raggedright\arraybackslash}m{0.25\textwidth}}
\caption{%
Representative software failures, deviations from the intended
experimental protocol, and gaps in traceability, together with the
checks or recorded evidence available in \fedbrew and when they are
assessed. A development-time check need not run during an ordinary experiment.}\label{tab:requirements}\\
\specialrule{1.6pt}{1.5pt}{1.5pt}
\textbf{Failure mode} &
\textbf{%
Condition checked or evidence recorded} &
\textbf{%
Check stage or evidence} \\
\specialrule{1.6pt}{1.5pt}{1.5pt}
\textbf{Ignored or incompatible configuration}\newline
{\footnotesize\itshape\color{metagray}the run silently proceeds on a default and
reports success}
&
An unrecognized configuration setting raises an error rather than falling back to default values
&
Configuration load
\\
\Line{table}{style=solid, color=burgundy, width=1.1pt}
\textbf{Invalid client partition}\newline
{\footnotesize\itshape\color{metagray}examples are duplicated or omitted,
changing client data and aggregation weights}
&
A generated partition is disjoint and exhaustive: no example is assigned to
two clients, and none is dropped.
&
Development-time check
\\
\Line{table}{style=solid, color=burgundy, width=1.1pt}
\textbf{Incorrect aggregation weights}\newline
{\footnotesize\itshape\color{metagray}
a run may finish with a weighted mean different from
the one required by its experimental protocol}
&
For classification, a client's aggregation weight is its full training-split
size, independently of how many examples its local update processes.
&
Development-time check
\\
\Line{table}{style=solid, color=burgundy, width=1.1pt}
\textbf{Non-finite client model state}\newline
{\footnotesize\itshape\color{metagray}the round completes, and the
corruption is folded into the aggregate and written into the next
checkpoint}
&
A non-finite client model state is rejected before contributing to the aggregate.
&
During aggregation
\\
\Line{table}{style=solid, color=burgundy, width=1.1pt}
\textbf{Pseudorandom seed reuse or inconsistent checkpoint recovery}\newline
{\footnotesize\itshape\color{metagray}the run completes although its pseudorandom streams or restored state differ from the intended experimental protocol}
&
Pseudorandom streams for client selection and data
loading are derived using the experiment seed and the relevant round,
client, and execution phase. %
Development-time checks compare interrupted and
uninterrupted runs under specified deterministic conditions; universal
bit-for-bit equivalence across execution modes or platforms is not claimed.
&
Development-time check
\\
\Line{table}{style=solid, color=burgundy, width=1.1pt}
\textbf{Untraceable result}\newline
{\footnotesize\itshape\color{metagray}the result cannot be traced to the experiment that produced it}
&
The run record stores its resolved configuration,
source-code revision when available, runtime metadata, and,
for manifest-based datasets, the hash of the dataset manifest. Metrics and checkpoints are periodically saved to disk at a configurable round interval. The manifest hash identifies its description,
not the contents of individual client data files; the recorded source-code
revision does not identify the contents of modified tracked or
untracked source files.
&
Run record (\texttt{run.json}), initialized
before training and updated when results are saved to disk
\\
\Line{table}{style=solid, color=burgundy, width=1.1pt}
\textbf{Uncontrolled extension}\newline
{\footnotesize\itshape\color{metagray}the extension runs to completion,
indistinguishable from a component that was checked}
&
An external component must register its name and configuration keys
before selection. Common configuration checks apply, but new
server--client combinations may require further software-compatibility
checks and scientific-condition checks supplied by their author.
&
Configuration load; additional development-time
checks when supplied
\\
\specialrule{1.6pt}{1.5pt}{1.5pt}
\end{longtable}
\endgroup

\subsection{Self-describing experiments}

The experiment configuration identifies the components and settings used
by the experiment implementation. Configuration loading rejects unknown
settings and invalid values, and rechecks the resolved configuration after
command-line overrides. Before training, the \emph{preflight} checks
supported combinations of the selected server strategy, client update rule,
task, model, dataset, evaluation, and execution settings. 
For manifest-based datasets, it checks the manifest and referenced files, client identities and counts, required evaluation splits, declared split sizes, and whether stored feature tensors match the manifest's declared data type and value range. 
For example, if a client data file contains floating-point features
but the dataset manifest declares integer-valued features
(\texttt{uint8}), the mismatch is detected before training begins.
The same preflight checks model-data shape compatibility and, for prepared language models, whether the required local asset manifest is usable, so a mistyped model-asset path fails before the job begins.
The \texttt{--validate-only} option runs the full preflight without
starting training.

During execution, \texttt{run.json} records the resolved experiment
configuration and available metadata. It is updated alongside the round
metrics whenever results are saved to disk, maintaining a record of
completed and interrupted runs.
The excerpt below, from the worked external-extension example
\texttt{examples/drift-quad}, illustrates three elements of the run record:
the recorded source-code revision, the dataset-manifest hash, and the
components registered by the extension.

\begin{lstlisting}[escapechar=|]
"code_state": {
  "git_commit": "bc800f5...8ff15b",       |{\color{metagray}// code}|
  "git_dirty": false
},
"dataset": {
  "manifest_sha256": "0b9d806...bd081f"   |{\color{metagray}// data}|
},
"extensions": {
  "examples/drift-quad/problem.py": {
    "registered": [                       |{\color{metagray}// added components}|
      ["generators", "drift_quad"],
      ["tasks", "drift_quad"],
      ["models", "quad_vector"]
    ]
  }
}
\end{lstlisting}

When Git version-control metadata is available, these fields record the exact commit and whether tracked files had uncommitted changes. 
A run outside a Git repository or without Git installed still proceeds and records why those fields are empty; \fedbrew does not modify the repository.
The recorded commit does not identify the contents of modified or untracked files, extension hashes identify the loaded extension files only, and the manifest hash does not cover individual shards.

Together, these records provide the configuration, source-code revision,
dataset description, and runtime context for interpreting and comparing
reported results.

%% file: sections/03-contracts.tex
\section{Checking Software and Scientific Conditions}
\label{sec:contracts}

The architecture in \Cref{sec:fedbrew-design} supports checks at
different stages of an experiment. Here, we describe how \fedbrew
connects specified scientific and software conditions to these checks
and examines whether the checks detect violations of those conditions.

A specified condition states what must hold, whereas a check tests
whether it holds for given inputs. For example, a scientific condition
may require that each training example is assigned to exactly one client.
A corresponding check examines the generated partition to determine
whether this condition holds; it has specified inputs, an execution stage,
and an observable result. Checks are performed at the stage relevant to
each condition. Before
training, configuration and preflight checks examine software conditions
such as recognized settings, manifest structure, and supported component
combinations. During execution, numerical checks reject non-finite client
states at aggregation. Development-time checks assess scientific conditions,
including partition disjointness and aggregation weights, against reference
cases. The run record preserves evidence for later inspection.

\paragraph{Configuration checks.}

\fedbrew uses closed configuration schemas:
\emph{an unrecognized setting is an error rather
than an invitation to use a default}. Checks also reject incompatible
settings before execution, and the resolved configuration is stored
with the run. For example, misspelling \texttt{learning\_rate} is rejected
instead of silently selecting its default value. The abbreviated diagnostic
is shown below:

\begin{commandblock}
\cmdline{cp configs/synthetic/fedavg.yaml /tmp/fedavg-typo.yaml}
\cmdline{sed -i 's/learning\_rate: 0.01/laerning\_rate: 0.01/' /tmp/fedavg-typo.yaml}
\cmdline{fedbrew run \opt{config} /tmp/fedavg-typo.yaml \opt{validate-only}}
\begingroup\catcode`\_=12\relax\raggedright\noindent
-- PREFLIGHT --\\
❙ ✗ load config              unknown configuration key: client.laerning_rate\\
1 error  0 warnings
\endgroup
\end{commandblock}

\paragraph{Data and evaluation checks.}

Development-time checks assess whether generated client partitions are
disjoint and exhaustive, and the resulting partition description is recorded
in a stable manifest. Before training, preflight checks the referenced
files, client and split counts, required evaluation splits, and the stored
tensors against their declared data type and value range.
The construction of training, validation, and test splits depends on
the dataset and its partitioning procedure. For example, the partitioned-test generator obtains each client's
test examples from the dataset's official test set, whereas FEMNIST
forms a separate test split from each writer's data. In both cases,
validation and test examples are disjoint. Metric names distinguish
which model is evaluated: \texttt{fit\_loss} evaluates each local model
before aggregation, while \texttt{train\_loss\_sample\_weighted\_avg}
evaluates the aggregated model on client training data.

\paragraph{Algorithm and numerical checks.}

Development-time checks compare aggregation and client update rule
implementations against expected results computed independently for
prescribed inputs (reference calculations). They also check
invariants, namely conditions that must hold for valid inputs and
executions. 
In particular, for classification, client aggregation weights use the full training-split sizes, even when the local updates process only a subset of those examples. 
The reference checks verify both the aggregation weights assigned to
participating clients and the resulting server aggregate against independently computed expected aggregation weights and the
corresponding sample-size-weighted average of the participating
clients' model states.
During aggregation, finite-value checks reject non-finite client model
states before the server update is committed.

The development-time checks in \path{tests/test_non_finite_aggregation.py}
supply client model states containing NaN or Inf values and verify that
aggregation rejects them. These checks also verify that the affected run is
recorded as diverged and no checkpoint is written for the invalid round.

Alongside these checks of model-update behavior, \fedbrew records
execution metrics for each simulated communication round. Communication
volume accounts for the model information exchanged in memory between
the server and clients, while computation metrics describe the work
performed under the selected execution mode. These records complement
the development-time reference checks and the runtime numerical
checks that reject non-finite client model states by characterizing how the updates
were executed.

\paragraph{Randomness and recovery checks.}

Client selection, data shuffling, and stochastic local updates use
pseudorandom streams whose behavior is part of the experimental protocol.
For example, a naive approach that initializes the client-selection generator with the experiment seed plus the round number can reuse the same seed in different experiments.
With this approach, experiment seed 42 at round 2 and experiment seed 43 at round 1 would both initialize the generator with seed 44, resulting in identical client selections when the client population and participation rate are unchanged.
To avoid this arithmetic coupling, \fedbrew derives deterministic seeds by hashing the experiment seed together with the relevant context, such as the communication round, client identity, and execution phase.

The resulting pseudorandom streams are reproducible and
context-specific. Data generation and training use separate seeds.
Checkpoints preserve the random-number generator states and algorithm
state needed to resume interrupted runs. Development-time checks compare
resumed and uninterrupted executions under specified deterministic
settings, while recovery checks reject incomplete restoration of coupled
server and client state. Older checkpoints lacking saved generator states
may resume with a warning.

\paragraph{Documentation consistency checks.}

Development-time checks compare algorithm registries with their
documentation in both directions. For example, removing a server
strategy from its registry while retaining it in the documentation
is reported as an inconsistency by \texttt{pytest}, the Python test
runner used in the repository. The diagnostic below illustrates
this mismatch:

\begin{commandblock}
\cmdline{python -m pytest tests/test\_docs\_algorithms.py -v}
\begingroup\catcode`\_=12\relax\raggedright\noindent
AssertionError: Items in the first set but not the second:\\
'fedlalr' : the chapter's server_strategies line and the registry disagree
\endgroup
\end{commandblock}

Conversely, the same check reports an implemented component that
is missing from the documentation.

To assess whether a check detects a specified violation, we
deliberately introduce invalid inputs or implementation mismatches,
such as unknown configuration settings and non-finite client model states.
Other checks examine partition disjointness and exhaustiveness, compare
aggregation with reference calculations, compare resumed and uninterrupted
runs, or check consistency between algorithm registries and documentation.
\Cref{tab:contract-validation} presents seven representative cases,
identifying the condition examined, expected response, and corresponding
check module. Configuration and runtime checks reject invalid operations
when a condition is violated, while development-time checks report
failures when an implementation disagrees with a reference calculation
or consistency condition. Together, these cases provide concrete
evidence of how the corresponding checks behave.

\begingroup
\small
\setlength{\tabcolsep}{4pt}
\renewcommand{\arraystretch}{1.0}
\begin{longtable}{
>{\raggedright\arraybackslash}m{0.29\textwidth}
>{\raggedright\arraybackslash}m{0.38\textwidth}
>{\raggedright\arraybackslash\footnotesize}m{0.26\textwidth}}
\caption{%
Representative checks using injected faults, reference calculations,
and consistency conditions, with their expected responses and corresponding
check modules.}\label{tab:contract-validation}\\
\specialrule{1.6pt}{1.5pt}{1.5pt}
\textbf{%
Check case} &
\textbf{Detection and stage} &
\textbf{%
Check module (\texttt{tests/})} \\
\specialrule{1.6pt}{1.5pt}{1.5pt}
\textbf{Unknown configuration key}\newline
{\footnotesize\itshape\color{metagray}Injected fault} &
Configuration check rejects the unknown key before training. &
\path{test_unknown_config_keys.py} \\
\Line{table}{style=solid, color=burgundy, width=1.1pt}
\textbf{Partition disjointness and exhaustiveness}\newline
{\footnotesize\itshape\color{metagray}Reference property} &
Development-time check compares client assignments with disjointness
and exhaustiveness conditions. &
\path{test_partition_disjointness.py} \\
\Line{table}{style=solid, color=burgundy, width=1.1pt}
\textbf{Manifest declares an incorrect tensor type or range}\newline
{\footnotesize\itshape\color{metagray}Injected fault} &
Preflight check reports a mismatch between declared and observed
tensor properties. &
\path{test_manifest_input_claims.py} \\
\Line{table}{style=solid, color=burgundy, width=1.1pt}
\textbf{Incorrect classification aggregation weight or weighted mean}\newline
{\footnotesize\itshape\color{metagray}Reference calculation} &
Development-time checks compare client training-split weights and the
server aggregate with reference values. &
\path{test_aggregation_weight_is_the_train_split_size.py};
\path{test_aggregation_correctness.py} \\
\Line{table}{style=solid, color=burgundy, width=1.1pt}
\textbf{Non-finite client model state}\newline
{\footnotesize\itshape\color{metagray}Injected fault} &
Aggregation rejects the non-finite update and does not checkpoint
the affected round. &
\path{test_non_finite_aggregation.py} \\
\Line{table}{style=solid, color=burgundy, width=1.1pt}
\textbf{Incomplete restoration of coupled server/client state}\newline
{\footnotesize\itshape\color{metagray}Injected fault} &
Recovery check rejects incomplete restoration of coupled server
and client state. &
\path{test_resume_is_all_or_nothing.py} \\
\Line{table}{style=solid, color=burgundy, width=1.1pt}
\textbf{Algorithm registry and documentation disagree}\newline
{\footnotesize\itshape\color{metagray}Consistency check} &
Development-time check detects differences between documented
algorithms and registry entries. &
\path{test_docs_algorithms.py} \\
\specialrule{1.6pt}{1.5pt}{1.5pt}
\end{longtable}
\endgroup

The listed modules can be run from the repository with
\texttt{python -m pytest}, allowing the reported behavior to be
inspected and reproduced.

%% file: sections/04-agent-ready.tex
\section{Agent-Ready Development}
\label{sec:agent-ready}

Here, \emph{agent-ready} describes a development workflow in which a coding agent works on a scoped task using shared component interfaces, relevant scientific conditions, and checks of specified software and scientific conditions.
The same resources guide human contributors.

\subsection{Repository knowledge for agents}

Repository guidance is organized into topic-specific chapters in
\texttt{docs/}. For example, \path{docs/07-algorithms.md} describes server
strategies and client update rules, while \path{docs/08-metrics.md}
defines the reported metrics. Each chapter ends with a \texttt{For agents}
section covering five items: relevant source paths, commands, invariants,
tests, and known failure modes. This structure helps researchers and coding
agents locate the implementation details and checks relevant to a change.

The implementation defines the technical behavior described by the
chapters. Some interface information is obtained directly from the code:
for example, \texttt{fedbrew --help} builds its subcommand list from
the \texttt{COMMANDS} mapping in \path{fedbrew/cli/dispatch.py}. 

For manually maintained documentation, automated checks compare selected
interface details with the implementation in both directions. For example,
\path{tests/test_docs_algorithms.py} compares server-strategy and client-rule
names in \path{docs/07-algorithms.md} with the corresponding registries.
It fails when a registered name is missing from the chapter or a documented
name is absent from the registry. Related checks cover configuration keys
and metric names, giving contributors direct feedback when interface changes
require documentation updates.

The root \texttt{README.md} serves as an overview and entry point to the
chapters. \path{tests/test_readme_is_an_overview.py} checks, among other
things, that it links to the documentation and leaves detailed configuration
keys and metric columns to their respective chapters. Keeping these details
in one reference location, with checks against the implementation, helps
maintain consistency as the codebase evolves.

\subsection{Extension workflow and checks}

To examine the extension workflow, we register new names for a server
strategy and a client update rule using an existing construction function,
without adding the corresponding declarations. This probe exercises the
checks for aggregation weights, checkpointed state, communicated state, and
configuration keys before a new algorithm is implemented. Seven
development-time test modules report failures and identify missing declarations. Selected failures are
shown below; \texttt{SUBFAILED} denotes a failed parametrized subtest.

\begin{lstlisting}
SUBFAILED(registry='server_strategies')
    RegistryListingTest::test_every_registered_name_is_listed
FAILED RegistryNamesTest::test_server_strategies_match_the_registry
FAILED ClassificationTest::test_every_registered_strategy_is_classified
FAILED MeterMatchesPayloadTest::test_every_registered_rule_has_a_builder
...
\end{lstlisting}

The failures identify specific follow-up tasks: document the new registry
entries, add a configuration that exercises the new rule, classify its
aggregation behavior, and define how its communication volume is accounted
for. Contributors can address these items and rerun the checks. Other
component types follow the same registration workflow with their relevant
validation checks.
\emph{Specified conditions already encoded in the repository become visible when the corresponding checks report violations.}
For scientific conditions not yet encoded, contributors specify them and implement corresponding checks. The repository can then assess them during development alongside existing checks.

%% file: sections/05-conclusion.tex
\section{Conclusion}
\label{sec:conclusion}

\fedbrew addresses protocol fidelity alongside computational
reproducibility by checking specified software and
scientific conditions of FL experiments. 
It combines configuration and data-manifest validation, numerical
checks during aggregation, development-time reference checks, and
self-describing run records. The checks assess specified conditions at the relevant stages;
checkpoints and recovery checks
support consistent resumption under specified deterministic settings. 
Together, these mechanisms provide targeted checks of the intended
experimental protocol and preserve the context needed to interpret reported results.
The explicit interfaces and registries also support new algorithms,
datasets, models, and tasks. These components reuse common configuration
and software-compatibility checks and artifact recording, while developers can
add checks for their specified scientific conditions. 
The same workflow serves researchers and coding agents: contributors
define the intended behavior and implement checks for its specified conditions. \fedbrew supports reproducible FL research through machine-checkable scientific conditions and inspectable
run records.

%% file: refs.bib
@inproceedings{karimireddy2020scaffold,
  title     = {{SCAFFOLD}: Stochastic Controlled Averaging for Federated Learning},
  author    = {Karimireddy, Sai Praneeth and Kale, Satyen and Mohri, Mehryar and Reddi, Sashank and Stich, Sebastian and Suresh, Ananda Theertha},
  booktitle = {Proceedings of the 37th International Conference on Machine Learning},
  series    = {PMLR},
  volume    = {119},
  pages     = {5132--5143},
  year      = {2020}
}

@inproceedings{hu2021lora,
  title     = {{LoRA}: Low-Rank Adaptation of Large Language Models},
  author    = {Hu, Edward J. and Shen, Yelong and Wallis, Phillip and Allen-Zhu, Zeyuan and Li, Yuanzhi and Wang, Shean and Wang, Lu and Chen, Weizhu},
  booktitle = {International Conference on Learning Representations},
  year      = {2022}
}

@article{beutel2020flower,
  title   = {Flower: A Friendly Federated Learning Research Framework},
  author  = {Beutel, Daniel J. and Topal, Taner and Mathur, Akhil and Qiu, Xinchi and Fernandez-Marques, Javier and Gao, Yan and Sani, Lorenzo and Li, Kwing Hei and Parcollet, Titouan and de Gusm{\~a}o, Pedro Porto Buarque and Lane, Nicholas D.},
  journal = {arXiv:2007.14390},
  year    = {2020}
}

@article{he2020fedml,
  title   = {{FedML}: A Research Library and Benchmark for Federated Machine Learning},
  author  = {He, Chaoyang and Li, Songze and So, Jinhyun and Zeng, Xiao and Zhang, Mi and Wang, Hongyi and Wang, Xiaoyang and Vepakomma, Praneeth and Singh, Abhishek and Qiu, Hang and Zhu, Xinghua and Wang, Jianzong and Shen, Li and Zhao, Peilin and Kang, Yan and Liu, Yang and Raskar, Ramesh and Yang, Qiang and Annavaram, Murali and Avestimehr, Salman},
  journal = {arXiv:2007.13518},
  year    = {2020}
}

@inproceedings{lai2022fedscale,
  title     = {{FedScale}: Benchmarking Model and System Performance of Federated Learning at Scale},
  author    = {Lai, Fan and Dai, Yinwei and Singapuram, Sanjay and Liu, Jiachen and Zhu, Xiangfeng and Madhyastha, Harsha and Chowdhury, Mosharaf},
  booktitle = {Proceedings of the 39th International Conference on Machine Learning},
  series    = {PMLR},
  volume    = {162},
  pages     = {11814--11827},
  year      = {2022}
}

@inproceedings{caldas2018leaf,
  title     = {{LEAF}: A Benchmark for Federated Settings},
  author    = {Caldas, Sebastian and Duddu, Sai Meher Karthik and Wu, Peter and Li, Tian and Kone{\v{c}}n{\'y}, Jakub and McMahan, H. Brendan and Smith, Virginia and Talwalkar, Ameet},
  booktitle = {Workshop on Federated Learning for Data Privacy and Confidentiality at NeurIPS},
  year      = {2019}
}

@article{meyer1992contract,
  title   = {Applying ``Design by Contract''},
  author  = {Meyer, Bertrand},
  journal = {Computer},
  volume  = {25},
  number  = {10},
  pages   = {40--51},
  year    = {1992},
  doi     = {10.1109/2.161279}
}

@inproceedings{claessen2000quickcheck,
  title     = {{QuickCheck}: A Lightweight Tool for Random Testing of {H}askell Programs},
  author    = {Claessen, Koen and Hughes, John},
  booktitle = {Proceedings of the Fifth ACM SIGPLAN International Conference on Functional Programming},
  pages     = {268--279},
  year      = {2000},
  doi       = {10.1145/351240.351266}
}

@article{segura2016metamorphic,
  title   = {A Survey on Metamorphic Testing},
  author  = {Segura, Sergio and Fraser, Gordon and S{\'a}nchez, Ana B. and Ruiz-Cort{\'e}s, Antonio},
  journal = {IEEE Transactions on Software Engineering},
  volume  = {42},
  number  = {9},
  pages   = {805--824},
  year    = {2016},
  doi     = {10.1109/TSE.2016.2532875}
}

@article{jia2011mutation,
  title   = {An Analysis and Survey of the Development of Mutation Testing},
  author  = {Jia, Yue and Harman, Mark},
  journal = {IEEE Transactions on Software Engineering},
  volume  = {37},
  number  = {5},
  pages   = {649--678},
  year    = {2011},
  doi     = {10.1109/TSE.2010.62}
}

@inproceedings{sculley2015technicaldebt,
  title     = {Hidden Technical Debt in Machine Learning Systems},
  author    = {Sculley, D. and Holt, Gary and Golovin, Daniel and Davydov, Eugene and Phillips, Todd and Ebner, Dietmar and Chaudhary, Vinay and Young, Michael and Crespo, Jean-Fran{\c{c}}ois and Dennison, Dan},
  booktitle = {Advances in Neural Information Processing Systems},
  volume    = {28},
  pages     = {2503--2511},
  year      = {2015}
}

@inproceedings{starace2025paperbench,
  title     = {{PaperBench}: Evaluating {AI}'s Ability to Replicate {AI} Research},
  author    = {Starace, Giulio and Jaffe, Oliver and Sherburn, Dane and Aung, James and Chan, Jun Shern and Maksin, Leon and Dias, Rachel and Mays, Evan and Kinsella, Benjamin and Thompson, Wyatt and Heidecke, Johannes and Glaese, Amelia and Patwardhan, Tejal},
  booktitle = {Proceedings of the 42nd International Conference on Machine Learning},
  series    = {PMLR},
  volume    = {267},
  pages     = {56843--56873},
  year      = {2025}
}

@inproceedings{hu2026reproagent,
  title     = {{ReproAgent}: Contract-Guided Paper-to-Code Reproduction},
  author    = {Hu, Xue and Pan, Zewei and Wang, Zhongyuan and Liu, Zhou and Su, Zeli and Zhang, Wentao},
  booktitle = {Findings of the Association for Computational Linguistics: EMNLP 2026},
  pages     = {3681--3702},
  year      = {2026},
  doi       = {10.18653/2026.findings-emnlp.175},
  previous_entry = {

@article{menzo2026harness,
  title   = {Reining in an Agentic Harness for High Energy Physics},
  author  = {Menzo, Tony and Fleming, George T. and Matchev, Konstantin T. and Mrenna, Stephen and Roman, Alexander},
  journal = {arXiv:2609.00107},
  year    = {2026}
}

@inproceedings{peimani2026esc,
    title={Executable Scientific Contracts for Verifiable {LLM} and Agent Research},
    author={Amir Reza Peimani},
    booktitle={NeurIPS 2026 Workshop on AI-Native Academia: Authorship, Peer Review, and Conference Governance under AI},
    year={2026},
    previous-entry = {@misc{peimani2026esc,
  title        = {Executable Scientific Contracts for Verifiable {LLM} and Agent Research},
  author       = {Peimani, Amir Reza},
  year         = {2026},
  version      = {1.1.1},
  howpublished = {Software release, GitHub repository},
  url          = {https://github.com/AmirRezaPeimani/executable-scientific-contracts},
  previous_entry = {

@article{polato2026fluke,
  title   = {Fluke: Federated Learning Utility Framework for Experimentation and Research},
  author  = {Polato, Mirko},
  journal = {Future Generation Computer Systems},
  volume  = {177},
  pages   = {108241},
  year    = {2026},
  doi     = {10.1016/j.future.2025.108241}
}

@article{wang2023flgo,
  title         = {{FLGo}: A Fully Customizable Federated Learning Platform},
  author        = {Wang, Zheng and Fan, Xiaoliang and Peng, Zhaopeng and Li, Xueheng and Yang, Ziqi and Feng, Mingkuan and Yang, Zhicheng and Liu, Xiao and Wang, Cheng},
  journal={arXiv:2306.12079},
  year={2023}
}

@article{xie2023federatedscope,
  title   = {{FederatedScope}: A Flexible Federated Learning Platform for Heterogeneity},
  author  = {Xie, Yuexiang and Wang, Zhen and Gao, Dawei and Chen, Daoyuan and Yao, Liuyi and Kuang, Weirui and Li, Yaliang and Ding, Bolin and Zhou, Jingren},
  journal = {Proceedings of the VLDB Endowment},
  volume  = {16},
  number  = {5},
  pages   = {1059--1072},
  year    = {2023},
  doi     = {10.14778/3579075.3579081}
}

@inproceedings{mcmahan2017communication,
  title={Communication-efficient learning of deep networks from decentralized data},
  author={McMahan, H. Brendan and Moore, Eider and Ramage, Daniel and Hampson, Seth and y Arcas, Blaise Ag\"uera},
  booktitle={Proceedings of the 20th International Conference on Artificial Intelligence and Statistics},
  pages={1273--1282},
  year={2017},
  volume={54}, 
  series={PMLR}
}
